\documentclass[a4paper,conference]{IEEEtran}
\ifCLASSINFOpdf

\else

\fi
\usepackage{amsmath,graphicx}
\usepackage{amsmath}
\usepackage{amssymb}
\usepackage{booktabs}
\usepackage{multirow}
\usepackage[vlined,ruled, linesnumbered]{algorithm2e}
\usepackage{balance}
\usepackage{subfigure}
\usepackage{balance}
\usepackage{url}
\usepackage{times}
\usepackage{epsfig}
\usepackage{graphicx}
\usepackage{amsmath}
\usepackage{amssymb}
\usepackage[table]{xcolor}

\begin{document}
\title{Stage-Wise Neural Architecture Search}

\author{\IEEEauthorblockN{Artur Jordao, Fernando Akio, Maiko Lie and William Robson Schwartz\\Smart Sense Laboratory, Computer Science Department\\Federal University of Minas Gerais, Brazil\\Email: \{arturjordao, fernandoakio, maikolie, william\}@dcc.ufmg.br}}

\maketitle

\begin{abstract}
Modern convolutional networks such as ResNet and NASNet have achieved state-of-the-art results in many computer vision applications. These architectures consist of stages, which are sets of layers that operate on representations in the same resolution. It has been demonstrated that increasing the number of layers in each stage improves the prediction ability of the network. However, the resulting architecture becomes computationally expensive in terms of floating point operations, memory requirements and inference time. Thus, significant human effort is necessary to evaluate different trade-offs between depth and performance. To handle this problem, recent works have proposed to automatically design high-performance architectures, mainly by means of neural architecture search (NAS). Current NAS strategies analyze a large set of possible candidate architectures and, hence, require vast computational resources and take many GPUs days. Motivated by this, we propose a NAS approach to efficiently design accurate and low-cost convolutional architectures and demonstrate that an efficient strategy for designing these architectures is to learn the depth stage-by-stage. For this purpose, our approach increases depth incrementally in each stage taking into account its importance, such that stages with low importance are kept shallow while stages with high importance become deeper. We conduct experiments on the CIFAR and different versions of ImageNet datasets, where we show that architectures discovered by our approach achieve better accuracy and efficiency than human-designed architectures. Additionally, we show that architectures discovered on CIFAR-10 can be successfully transferred to large datasets. Compared to previous NAS approaches, our method is substantially more efficient, as it evaluates one order of magnitude fewer models and yields architectures on par with the state-of-the-art.
\end{abstract}
\section{Introduction}\label{sec:introduction}
Convolutional networks have led to a series of breakthroughs in different tasks, pushing the state-of-the-art in many applications~\cite{He:2016, Tan:2019:ICML, Strubell:2019}. 
According to previous works~\cite{He:2016, Huang:2017, Tan:2019:ICML}, network depth is a major determinant factor for these achievements. While the prediction ability of very deep convolutional networks is impressive, often surpassing human performance~\cite{He:2015}, the resulting computational cost hinders applicability on low-power and resource-constrained devices.
%
\begin{figure}[!t]
	\centering
	\includegraphics[width=\linewidth]{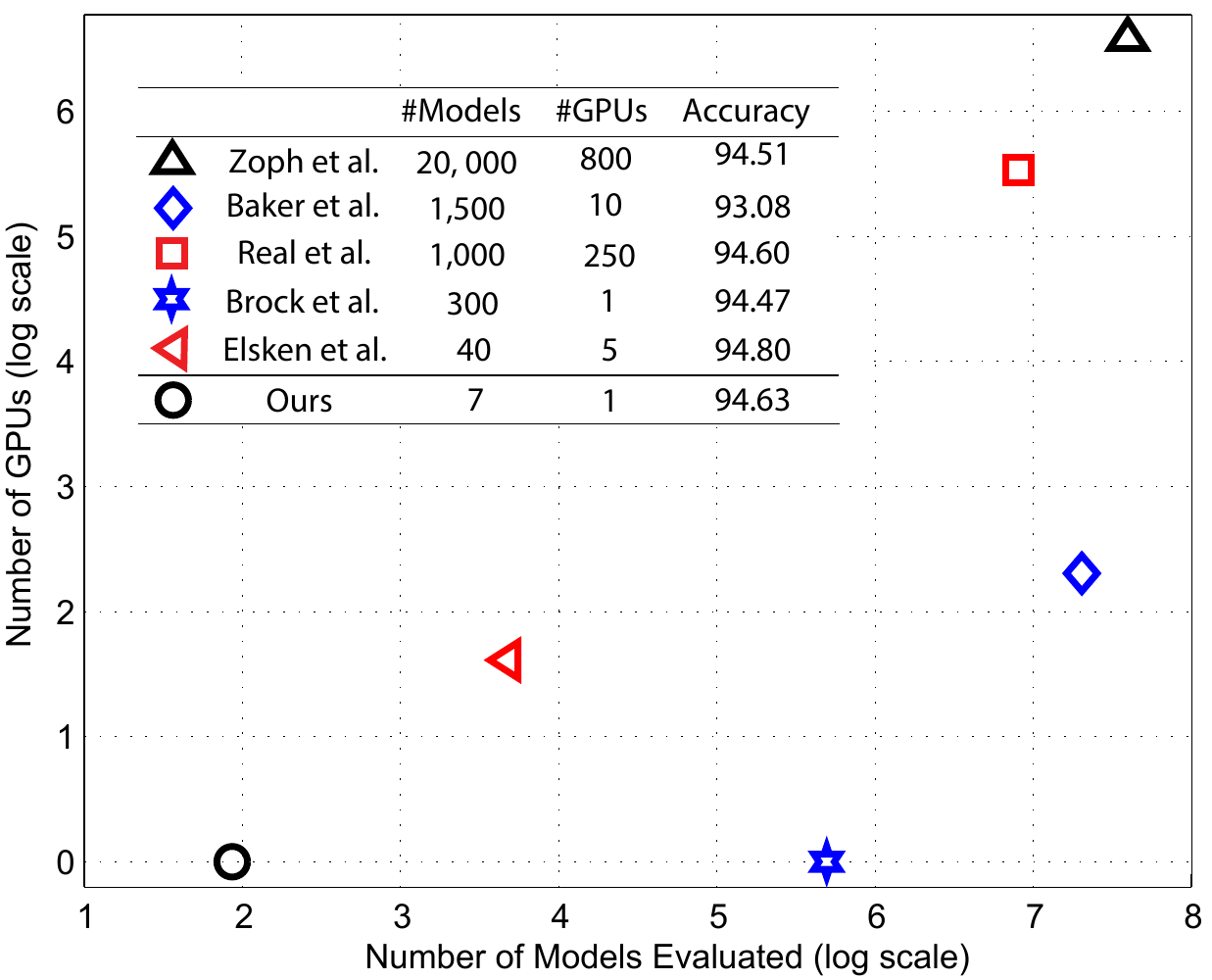}
	\caption{Comparison with state-of-the-art NAS. Different from previous works, instead of analyzing a large pool of architectures, we iteratively increment architectures taking into account the importance of the components (layers) to be inserted. As a consequence, our method discovers competitive and high-performance architectures by exploring one order of magnitude fewer models compared to other approaches. In addition, our method is the most resource-efficient as it designs architectures in a few hours on a single GPU.}
	\label{fig:teaser}
\end{figure}

\begin{figure}[!t]
	\centering
	{\includegraphics[width=0.8\columnwidth]{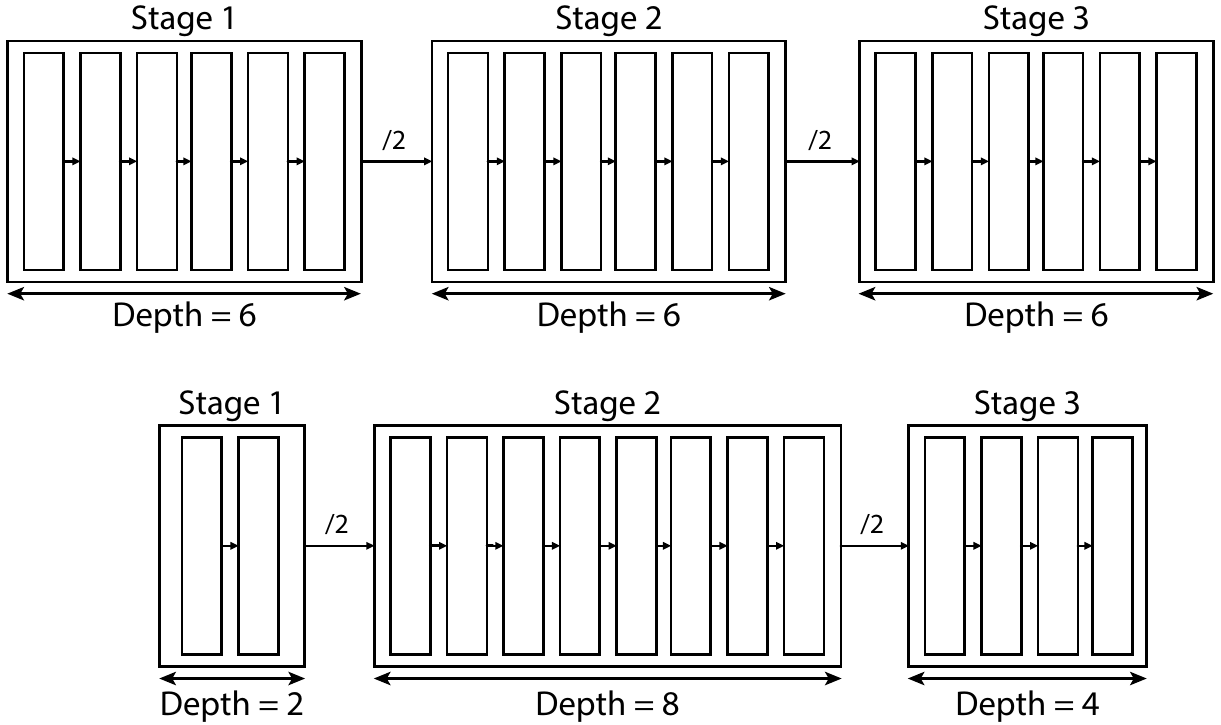}}
	\caption{\textbf{Top:} Structure of modern architectures, in which depth (number of layers) is the same for all stages. \textbf{Bottom:} Structure of our architectures, in which the depth of each stage is adjusted based on the importance of its features. In this example, the mid-stage of our architecture is more important as it is deeper, while the early-stage is less important as it is the shallower.}
	%
	\label{fig:architectures}
\end{figure}

To handle the problem above, it is possible to evaluate different trade-offs between accuracy and network complexity (e.g., number of filters and layers)~\cite{Howard:2017, Sandler:2018,Wang:2018}, however, this process requires significant human engineering due to its trial-and-error essence. 
Neural architecture search (NAS) alleviates this problem by learning to design architectures automatically. Given a criterion such as accuracy or fixed resource budget, NAS explores many possible candidate architectures to optimize the target criterion. Due to the large number of evaluated architectures, most NAS approaches require vast computational resources, parallel processing infrastructure and take many days to process even with modern GPUs~\cite{Gordon:2018, Jin:2019, Dong:2019} (see Figure~\ref{fig:teaser}). For example, the approach by Real et al.~\cite{Real:2017} requires $10$ days on $250$ GPUs to design architectures competitive with the state-of-the-art.
Additionally, architectures discovered by NAS are, generally, very complex and may lack efficient implementation in current deep learning frameworks~\cite{Wang:2018, Chen:2019}.
In this work, we propose a simple, effective, and efficient method for automatic design of convolutional neural networks. We note the fact that modern architectures are composed of stages --- groups of layers operating on the same resolution, and that in popular architectures the depth of these stages is defined either uniformly (e.g., ResNet39--110) or empirically (e.g., ResNet50--101) despite experiments showing that some stages are more important to the generalization ability of the network than others~\cite{Huang:2018, Wang:2018:ECCV, Greff:2017}. For example, Huang et al.~\cite{Huang:2018} noticed that accuracy degrades more by removing filters from early stages than from deeper stages, while Greff et al.~\cite{Greff:2017} and Wang et al.~\cite{Wang:2018:ECCV} showed that some stages could be shorter than others with negligible impact on accuracy. Motivated by these findings, our hypothesis is that we can adjust depth on a stage-by-stage basis to build shallow, hence low-cost, networks with the same predictive ability (or better) than deeper networks. To demonstrate this, we propose a method that increases the depth of stages iteratively by evaluating the importance of the features they output. Stages with low importance are kept shallow, while stages with high importance have their depth increased since they are more likely to improve the generalization ability of the network (Figure~\ref{fig:architectures}). We assess our approach using three different feature importance estimation methods, namely Partial Least Squares~\cite{Geladi:1986}, Infinite Feature Selection~\cite{Roffo:ICCV} and Infinite Latent Feature Selection~\cite{Roffo:2017}, and show that our method is effective with all of them. It is worth noting that, while our strategy does not optimize computational cost directly, the architectures it designs are efficient since their depth is not increased unnecessarily.

The contributions of this work are the following: 
(i) We present a simple, effective, and efficient approach for automatic design of compact convolutional networks;
(ii) We demonstrate that shallower networks can achieve similar or even better accuracy than deeper networks when the depth is set according to the importance of features output across the network. Similar to cardinality~\cite{CVPR:2017}, this finding enables increasing the capacity of networks without compromising its efficiency;
(iii) We show that the discovered architecture on small datasets can be successfully transferred to large datasets.

Although other components (i.e., number of filters) could be learned stage-by-stage, recent works have demonstrated that optimizing the number of layers leads to more efficient convolutional networks~\cite{Tan:2019:ICML}. Thus, by exploring only depth, our search space is small, hence, computationally efficient. Specifically, while state-of-the-art NAS approaches evaluate hundreds or even thousands of models, our method considers less than ten models to achieve competitive results, as illustrated in Figure~\ref{fig:teaser}. 
Our approach is supported by the unraveled view of residual networks~\cite{Veit:2016, Greff:2017}, which predicts that each stage in the network learns a single level of representation and that layers from a single stage only refine representations on the same level. Since there is no reason to assume that all levels of representation require the same level of refinement to perform effectively on a given task, it is reasonable to learn directly from data how deep each stage should be.
To further improve the performance of our method, we also propose a version that reuses weights of existing and pre-trained convolutional networks, reducing even more the computation time required for discovering architectures.
%

To assess the quality of our architectures, we conduct experiments on the CIFAR-10~\cite{krizhevsky2009}, Tiny ImageNet~\cite{Le:2015} and ImageNet datasets. For the ImageNet dataset, besides its original version ($224\times224$ images)~\cite{ilsvrc15}, we also consider its $32\times32$ version~\cite{Chrabaszcz:2017}.
On these datasets, compared to deeper networks, our discovered architectures achieve better accuracy with up to $43.63\%$ and $41.10\%$ less memory consumption and FLOPs, respectively.
When considering other candidate architectures, these results are consistent even taking into account shallower networks (e.g., ResNet44).
%
Following a recent trend~\cite{Lacoste:2019, Schwartz:2019, Strubell:2019}, we also measure the carbon emission for training our architectures, showing that our approach emits up to $41\%$ less carbon compared to human-designed networks.
Compared to NAS strategies, our method is extremely more efficient, as it evaluates one order of magnitude fewer models and discovers architectures on par with the state-of-the-art. The source code and models are available at: \emph{https://github.com/arturjordao/StageWiseArchitectureSearch}
\section{Related Work}\label{sec:related_work}
While convolutional networks are now the standard in visual recognition tasks, these networks generally rely on design by human experts to achieve state-of-the-art effectiveness. Consequently, there have been substantial efforts to automate the process of creating such networks. 
To this end, a typical technique is to use reinforcement learning (RL) to generate candidate architectures.
Backer et al.~\cite{Baker:2017} employed this strategy for selecting types of layers and their parameters (i.e., depth, receptive field, stride).
In contrast, Zoph et al.~\cite{Zoph:2018} proposed to learn transferable architectures by applying the scheme of human-designed convolutional networks, in which layers share a similar structure. Their method uses a recurrent neural network to predict a cell, which consists of a set of layers (e.g., convolution, identity, pooling) and their connections. The final architecture is obtained by repeating the best cell $N$ times, where $N$ is manually predefined. 
Similarly to Zoph et al.~\cite{Zoph:2018}, we show that our method is capable of building architectures that generalize well across datasets, such that we can learn a model on a small dataset and transfer it to large datasets. In addition, our method is orthogonal to this approach in the sense that we discover $N$ given a predefined cell.

Since using neural networks to learn architectures is time-consuming and require careful parameter setting~\cite{Wu:2018, Dong:2019}, many works employ evolutionary algorithms to guide the search~\cite{Real:2017, Dong:2020}. 
For example, Real et al.~\cite{Real:2017} used an evolutionary framework to build convolutional networks, in which each candidate architecture is an individual of the population and the mutation stage consists of operations such as inserting or removing layers/connections.

Although RL and evolutionary NAS are capable of building accurate models, their search process is computationally expensive since each candidate architecture needs to be trained from scratch. To handle this problem, recent works attempt to transfer the knowledge of previous pre-trained networks to the candidate architectures~\cite{Elsken:2018, Kandasamy:2018, Jin:2019}.
For this purpose, a popular technique is network morphism, which creates new networks by means of function-preserving transformations~\cite{Chen:2016}. In essence, network morphism allows the original and the modified network to have the same prediction ability.
Elsken et al.~\cite{Elsken:2018} employed network morphism to initialize architectures, aiming at reducing the cost of training them from scratch. 
%
Cai et al.~\cite{Cai:2018} applied RL to generate transformations on an initial network, for example, DenseNet. As suggested in their work, using an existing and pre-trained architecture is an efficient manner of exploring the search space, being possible to reuse its weights as well as its successful initial structure. Our method takes advantage of these observations, but focuses exclusively on depth.
Kandasamy et al.~\cite{Kandasamy:2018} and Jin et al.~\cite{Jin:2019} employed Bayesian optimization to guide transformations during the search process. While computationally efficient, these approaches yield low-accuracy architectures. To achieve competitive results, many hyper-parameters need to be set manually~\cite{Jin:2019}, rendering an unfair comparison with other NAS approaches. Compared to morphism-based NAS, our method enables reusing weights of pre-trained convolutional networks more easily because it does not require a careful selection of the morphism operations.

Current NAS approaches need to train or fine-tune each candidate architecture. Since these approaches yield hundreds or even thousands of candidates (see Figure~\ref{fig:teaser}), the search process is costly and prohibitive on large datasets. To alleviate this problem, most NAS approaches train the candidate architectures for few epochs, which might yield unreliable models during the search process~\cite{Dong:2020, Sciuto:2020}.
Our method, on the other hand, evaluates substantially fewer models enabling us to train them for more epochs. In particular, on the CIFAR-10 dataset, our method achieves accuracy competitive with state-of-the-art NAS methods while evaluating only seven models, one order of magnitude less than these approaches.
%
%

%

Concurrently to our work, the approach by Dong et al.~\cite{Dong:2019} discovers accurate architectures efficiently. However, their method fails to design architectures when applied directly on large datasets since their approach needs careful tuning and different hyper-parameters on these datasets.
In contrast, our search leads to competitive architectures on both small and large datasets without requiring careful parameter setting.
\section{Proposed Method}\label{sec:proposed_method}

\begin{figure}[!t]
	\centering
	\includegraphics[width=\columnwidth]{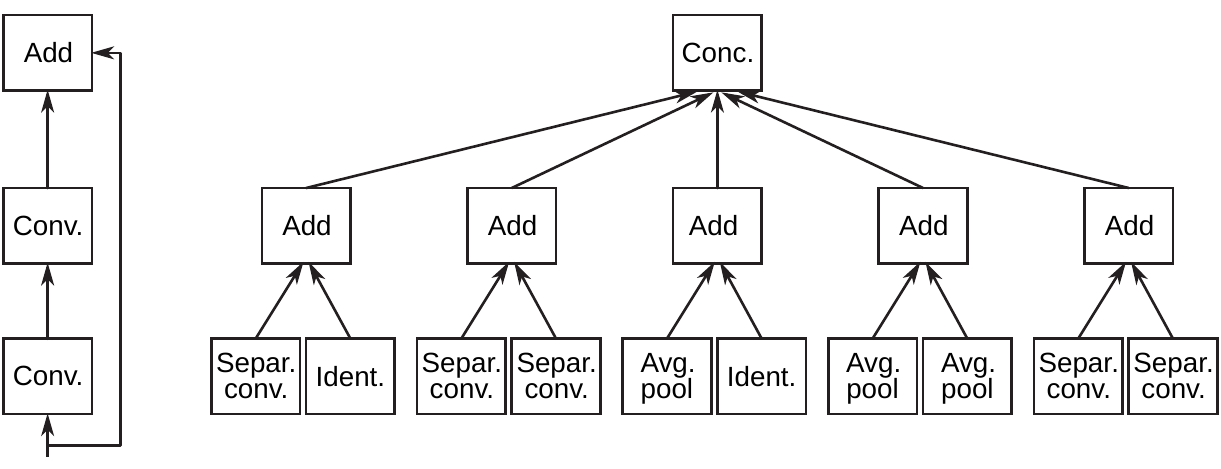}
	\caption{\textbf{Left:} Residual modules employed in ResNets~\cite{He:2016}. \textbf{Right:} Cell modules employed in NASNets~\cite{Zoph:2018}.}
	\label{fig:cell}
\end{figure}

\noindent
\textbf{Problem Statement.} 
Let $A$ be a convolutional network composed of $S$ stages. Each stage $s_i$ consists of $m_i$ modules (set of layers as illustrated in Figure~\ref{fig:cell}), which can be residual blocks or cells from ResNet and NASNet, respectively.
Following the structure of modern architectures, the layers within a stage operate on the same input/output resolution (i.e., their feature maps have the same dimension). When moving from a stage to another, a downsampling layer (indicated as $/2$ in Figure~\ref{fig:architectures}) is applied to reduce the spatial resolution. 

In previous works, including NAS, $m$ is the same for all stages or defined empirically. 
For instance, ResNet39 has six residual blocks in each of its stages (i.e., $m_{i \in \{1,..., S\}}=6$), as shown in Figure~\ref{fig:architectures} (top).
Recent works have demonstrated that some stages have a higher influence on the prediction ability of the network than others~\cite{Veit:2016, Greff:2017, Huang:2018}, suggesting that the number of modules within stages should be adjusted based on its importance. Motivated by this, we propose to learn the number of modules $m_i$ for each stage $s_i$ based on the importance of the features it outputs. 
We define a module as being residual blocks from ResNet (Figure~\ref{fig:cell}, left) or cells from NASNet (Figure~\ref{fig:cell}, right). We do not explore the combination of both, meaning that the discovered architecture is either ResNet-based or NASNet-based. We limit our experiments to these two types of modules due to their relevance in modern architectures and because our approach is grounded on the unraveled view of residual networks~\cite{Veit:2016}, for which there is only evidence for modules with skip-connections.

While other parameters could be learned stage-by-stage, adjusting depth is what provides the higher improvement in accuracy with the lower increase in computational cost~\cite{Tan:2019:ICML}. 
Therefore, we explore only depth since we are searching for accurate yet efficient networks.

\noindent
\textbf{Importance Score.}
To measure the importance score of a stage, we apply the process suggested by Jordao et al.~\cite{Jordao:2020}, which is the following. Given a stage $s_i$ of a convolutional network, we present the training samples to the network and extract the feature maps from the last layer of this stage. The reason for considering the last layer is that, as demonstrated by previous works~\cite{Veit:2016, Huang:2017, Greff:2017}, it contains information about previous layers, hence, about the entire stage. It is important to mention that this claim is valid only when the identity (i.e., skip-connection layer) is propagated to successive layers~\cite{Veit:2016}. 
%
%

Let $X_i$ be the features of $s_i$ estimated following the procedure above. The next step is to compute the importance of these features and average their values to compose the final importance score for each stage~\cite{Jordao:2020}. Specifically, by estimating the importance of $X_i$ we are estimating the importance of the stage $s_i$. 
We perform experiments using three different feature selection techniques, namely Partial Least Squares, Infinite Feature Selection, and Latent Infinite Feature Selection, which we describe briefly\footnote{Due to lack of space, we recommend the works by Geladi et al.~\cite{Geladi:1986}, Mehmood et al.~\cite{Mehmood:2012} and Roffo et al.~\cite{Roffo:ICCV, Roffo:2017} for more details.} below.

Partial Least Squares (PLS)~\cite{Geladi:1986} is a dimensionality reduction that projects a high-dimensional space onto a low-dimensional space such that the projected data has maximum covariance with its labels. To operate as a feature selection method, a technique named Variable Importance in Projection~\cite{Mehmood:2012} is applied to compute the contribution of each feature of high dimensional in generating the low-dimensional space.

Infinity Feature Selection (inf-FS)~\cite{Roffo:ICCV} is an unsupervised feature selection method that interprets features as vertices in an indirected fully-connected graph. The edges in this graph model the pairwise relation between features. Under this model, a path of length $l$ represents a subset of $l$ features and the idea is to expand $l$ to infinity using convergence property of the geometric power series of matrices. 
Improving upon this model, a supervised version of inf-FS (Infinite Latent Feature Selection~\cite{Roffo:2017} – il-FS) performs discriminative quantization of the raw features into a small set of tokens (that will be the edges) before creating the graph and expanding $l$ to infinity.
%
%

While PLS captures the relationship between features and their labels, inf-FS considers only the variance and the correlation between features. By using these approaches we measure the stage importance taking into account different aspects.
Although other feature selection methods could be employed, PLS and inf-FS approaches have presented impressive results in describing the importance of components from convolutional networks~\cite{Yu:2018, Jordao:2020}. In addition, PLS and inf-FS are robust to the number of samples, making our method suitable to small datasets as well as scalable for large datasets (in terms of memory, e.g., using few data samples).

\noindent
\textbf{Adjusting Stage Depth.}
Once we are able to estimate the importance score $\alpha_i$ for each stage $s_i$, the next step is to build a candidate architecture by adjusting the depth of each stage based on its importance. To this end, we first create a network $A$ with $S$ stages ($|s|=S$), each one containing the same number of modules, for example, by employing $S=3$ and $m_{i \in \{1, ..., S\}} = 6$ (i.e., ResNet39 in Figure~\ref{fig:architectures}, top).
Then, we create a temporary architecture $T$ by increasing the depth of $s_i$ to $m_i + \delta$, where $\delta$ is the growth step, i.e., the number of modules that can be inserted in a stage in a single iteration. 
%
Afterward, we compute the importance scores $\alpha_{A,i}$ and $\alpha_{T,i}$, for each stage $s_{i}$ of the initial and temporary architectures, respectively.
Finally, we update $m_i$ to $m_i + \delta$ if $\alpha_{T,i} > \alpha_{A,i}$ and create a candidate architecture $\hat{A}$ using the updated $m_i$.  It is worth mentioning that the importance scores are comparable in terms of magnitude.
The idea behind this incremental process is to measure if increasing depth will improve the representation learned by the candidate architecture.

The process above composes one iteration of our method, where at the end of each iteration one candidate architecture is discovered. The input for the next iteration is the candidate architecture designed with the values of $m_i$ updated. Algorithm~\ref{alg::auto_ml} summarizes all the steps of the proposed method.

The novelty of our method is that instead of sampling many architectures varying the $m_{th}$ values and using reinforcement learning or similar strategies to filter the best candidate, we generate one candidate architecture at a time iteratively.
%
In practice, given $n$ iterations, our method creates only $2n+1$ architectures, which is an order of magnitude fewer than state-of-the-art NAS approaches.

\noindent
\textbf{Weight Transfer Technique.}
Similar to previous NAS approaches~\cite{Real:2017, Zoph:2018}, the process of creating a model consists of training it from scratch for some epochs, which can be computationally prohibitive for large datasets such as ImageNet.
However, since our method employs the same structure (i.e., modules) of existing architectures, we propose to transfer the knowledge (weights) from a pre-trained network to our candidate architecture. For example, when employing residual modules, our candidate architecture can use the weights of a pre-trained ResNet. 
This way, instead of training from scratch, we only need to adjust the weights by fine-tuning for a few epochs to compensate changes in the magnitude of the feature maps~\cite{Veit:2016, Greff:2017}. 
One restriction of this strategy is that the depth of a stage of the candidate architecture cannot exceed the depth of the network that is providing the weights.
In practice, we show that this does not occur as our candidate architectures are shallower than existing networks.
In essence, our weight transfer technique is similar to the morphism strategy, but, this solution is simpler since it does not require careful selection of the morphism operations~\cite{Cai:2018, Jin:2019}.
\begin{algorithm}[!t]
\SetAlgoLined
\DontPrintSemicolon
\caption{Stage-Wise Neural Architecture Search}
\label{alg::auto_ml}
\SetKwInOut{Input}{Input}
\SetKwInOut{Output}{Output}
\Input{Number of iterations $n$, Number of stages $S$\\
	\hspace{0.1em}Initial number of modules per stage $m_0$\\
	\hspace{0.1em}Growth step $\delta$\\}
\Output{Set of candidate architectures $\mathbb{C}$}
\BlankLine
\BlankLine
Create $A$ with $S$ stages and $m_0$ modules each\\
\For{$k \leftarrow 1$ \KwTo $n$}{
	Create $T$ with $S$ stages and $m_i + \delta$ modules each\\
	\For{$i \leftarrow 1$ \KwTo $S$}{
		Compute importance scores $\alpha_{A,i}$ and $\alpha_{T,i}$\\
		\If {$\alpha_{T,i} > \alpha_{A,i}$}{
			$m_i \leftarrow m_i + \delta$\\}
	}
	Create $\hat{A}$ with $S$ stages and the updated $m_i$\\
	$A \leftarrow \hat{A}$\\
	$\mathbb{C} \leftarrow \mathbb{C} \cup \{\hat{A}\}$
}
\end{algorithm}
\section{Experiments}\label{sec:experiments}

\noindent
\textbf{Experimental Setup.}
We conduct experiments using a single NVIDIA GeForce GTX 1080 Ti GPU on a machine with 64 GB RAM. 
%
In our experiments, training from scratch consists in training the architectures for $200$ epochs applying horizontal random flip and random crop data augmentation. We employ SGD with a learning rate starting at $0.01$ and decreased by a factor of $10$ on the $100$ and $150$ epochs. When using the proposed weight transfer technique, the architectures are fine-tuned for $50$ epochs.
For fairness with previous works, which adjust their final architecture using additional epochs~\cite{Dong:2019, Brock:2018, Elsken:2018}, at the end of each iteration of Algorithm~\ref{alg::auto_ml} the candidate architecture is further trained for $100$ epochs.

To set the parameters of PLS, inf-FS and il-FS, we use a validation set from CIFAR-10 and select the parameters which led to the architecture with highest accuracy after running one iteration of Algorithm~\ref{alg::auto_ml}.
The growth step $\delta$ was set as $2$ and $1$ when using residual and cell modules, respectively. Although we have not exhaustively searched this parameter, we found that using these values lead to architectures with a good trade-off between accuracy and computational cost. Once we set these parameters, we apply them, unchanged, to other datasets.
%

Following previous works~\cite{Zoph:2018,Dong:2019, Dong:2020}, for low-resolution datasets (i.e., CIFAR-10, Tiny ImageNet and ImageNet $32\times32$ version) we set $S=3$. For the original version of the ImageNet dataset (images with $224\times224$ pixels), we set $S=4$, as suggested by He et al.~\cite{He:2016}. These values of $S$ allow a fair comparison with human-designed architectures as well as the NAS approaches. In addition, keeping $S$ fixed, we are able to measure only the influence of the depth of each stage.
\begin{table}[!b]
	\centering
	\renewcommand{\arraystretch}{1.2}
	\caption{Influence of the initial number of modules $m_0$ on the first candidate architecture.}
	\label{tab:k_values}
	\begin{tabular}{cccccc}
		\hline
		$m_0$ & Depth & \begin{tabular}[c]{@{}c@{}}Parameters\\ (Million)\end{tabular} & \begin{tabular}[c]{@{}c@{}}FLOPs\\ (Million)\end{tabular} & \begin{tabular}[c]{@{}c@{}}Memory\\  (MB)\end{tabular} & \begin{tabular}[c]{@{}c@{}}Accuracy\\  (200 epochs)\end{tabular} \\ \hline
		2 & 23 & 0.36 & 45 & 3.81 & 91.23 \\
		4 & 31 & 0.40 & 64 & 5.31 & 91.57 \\ 
		6 & 43 & 0.60 & 92 & 7.58 & 92.03 \\
		8 & 63 & 0.95 & 139 & 11.56 & 91.98 \\
		10 & 67 & 1.10 & 149 & 12.34 & 92.12 \\ \hline
	\end{tabular}
\end{table}
\noindent
\textbf{Influence of Initial Depth}. Our first experiment evaluates the influence of the depth $m_0$ of the initial architecture ($A$ in step $1$ of Algorithm~\ref{alg::auto_ml}). 
To this end, we vary $m_0$ from $2$ to $10$ in steps of $2$ and measure the performance of the resulting architecture after running one iteration of our method.

According to Table~\ref{tab:k_values}, we observe that large values of $m_0$ lead to accurate architectures, but the computational cost increases substantially as well. For example, for $m_0=10$ the candidate architecture after one iteration of Algorithm~\ref{alg::auto_ml} achieves an accuracy of $92.12$ with $1.10$ million parameters and $149$ million FLOPs. With $m_0=6$, on the other hand, the first candidate architecture obtains an accuracy of $92.03$ leading to significantly fewer parameters and FLOPs.
Note that this behavior also occurs in residual networks. For example, ResNet56 ($m=9$) is only $0.2$ percentage points (p.p) more accurate than ResNet44 ($m=7$), see Table~\ref{tab:comparison_scratch_ws_resnet}.

Based on Table~\ref{tab:k_values}, the best trade-off between accuracy and computational cost is achieved with $m=6$. Thus, we employ this value in the remaining experiments.
%
%
%
%

\noindent
\textbf{Importance Criteria.} This experiment assesses the quality of the candidate architectures discovered applying PLS and inf-FS to measure the importance of the stages. 
%

According to Table~\ref{tab:importance_criteria}, the proposed method using PLS designs more accurate candidate architectures. 
%
While the superiority of PLS could be attributed at first to the fact that it is supervised, in contrast to inf-FS, we also assessed a supervised variant of inf-FS, il-FS, and observed the same trend (Table~\ref{tab:importance_criteria}).
This suggests that PLS is more suitable for measuring the importance of the stages.
\begin{table}[!t]
	\centering
	\renewcommand{\arraystretch}{1.2}
	\caption{Accuracy on CIFAR-10 (200 epochs) of our method using different criteria for determining stage importance.}
	\label{tab:importance_criteria}
	\begin{tabular}{cccccc}
		\hline
		& \multicolumn{5}{c}{Iteration} \\ \cline{2-6} 
		Criterion & 1 & 2 & 3 & 4 & 5 \\ \hline
		inf-FS~\cite{Roffo:ICCV} & 91.59 & 92.09 & 92.02 & 92.36 & 92.45 \\
		il-FS~\cite{Roffo:2017} & 91.94 & 92.06 & 92.10 & 92.08 & 92.52 \\
		PLS~\cite{Geladi:1986} & 92.03 & 92.38 & 92.62 & 92.53 & 92.58 \\ \hline
	\end{tabular}
\end{table}

\begin{figure}[!b]
	\centering
	\includegraphics[width=\linewidth]{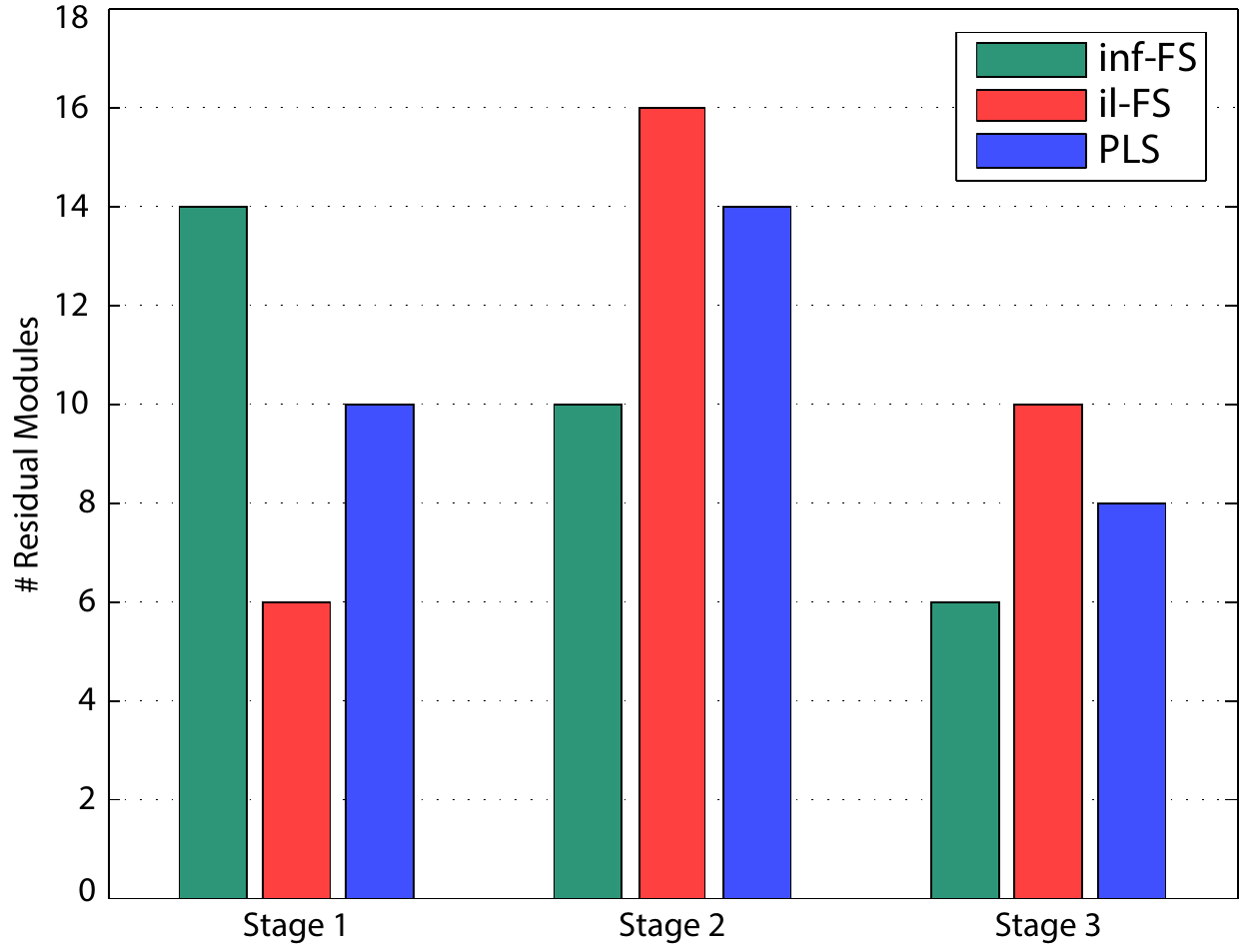}
	\caption{Number of residual modules, per stage (i.e. $m_i$), after five iterations of the proposed method using different criteria for determining stage importance.}
	\label{fig:residual_distribution}
\end{figure}
Figure~\ref{fig:residual_distribution} shows the distribution of modules resulting from different criteria for measuring the importance of architecture stages.
The results show that the approach used to measure importance has significant impact on the final architecture. 
%
In addition, our method applying both PLS and il-FS inserted more modules on the middle stage, which means increasing its depth brings improvements to the architecture. Importantly, this behavior is consistent with the work by Wang et al.~\cite{Wang:2018:ECCV}, where they demonstrate that removing layers from the middle stage degrades accuracy more than other stages.
%
%
This supports the fact that our approach is capable of identifying stages that need become deeper (most important) and the ones that could be kept shallow (least important).
\begin{table*}[!!ht]
	\centering
	\renewcommand{\arraystretch}{1.2}
	\caption{Comparison with human-designed architectures. Our architectures achieve superior accuracy and are more efficient. kgCO$_2$eq indicates carbon emission during the training step (the lower the better).}
	\label{tab:comparison_scratch_ws_resnet}
	\begin{tabular}{cccccccc}
		\hline
		Architecture & Depth & \begin{tabular}[c]{@{}c@{}}Param.\\ (Million)\end{tabular} & \begin{tabular}[c]{@{}c@{}}FLOPs\\ (Million)\end{tabular} & \begin{tabular}[c]{@{}c@{}}Memory\\ (MB)\end{tabular} & \begin{tabular}[c]{@{}c@{}}CPU Inference Time\\ (Milliseconds)\end{tabular} & \begin{tabular}[c]{@{}c@{}}Carbon Emission\\ (kgCO$_2$eq)\end{tabular} & \begin{tabular}[c]{@{}c@{}}Accuracy CIFAR-10 \\ (300 epochs)\end{tabular} \\ \hline
		ResNet44 & 44 & 0.66 & 97 & 8.14 & 36.33 & 0.96  & 92.83 \\
		Ours (it=1), scratch & 43 & 0.60 & 92 & 7.41 & 36.26 & 0.94 & 93.38 \\
		Ours (it=1), weight transfer & 39 & 0.56 & 83 & 7.00 & 31.69 & 0.20 & 93.32 \\ \hline
		ResNet56 & 56 & 0.86 & 125 & 10.42 & 46.86 & 1.27 & 93.03 \\
		Ours (it=3), scratch & 59 & 0.69 & 130 & 10.32 & 50.16 & 1.23 & 93.36 \\
		Ours (it=3), weight transfer & 51 & 0.90 & 111 & 9.16 & 40.96 & 0.24 & 93.61 \\ \hline
		ResNet110 & 110 & 1.7 & 253 & 20.67 & 90.33 & 2.26 & 93.57 \\
		Ours (it=5), scratch & 67 & 0.88 & 149 & 11.65 & 57.96 & 1.32 & 94.27 \\
		Ours (it=5), weight transfer & 59 & 1.23 & 139 & 11.31 & 49.99 & 0.29 & 93.51 \\ \hline
	\end{tabular}
\end{table*}

Based on these results, in the next experiments, we employ only PLS to measure the importance scores. Moreover, we report results considering one, three and five iterations and $300$ epochs of training. We observe that increasing the number of iterations above five does not improve accuracy substantially enough to justify the increase in computational cost.
%
%

\noindent
\textbf{Weight Transfer.}
Our next experiment evaluates the behavior of the proposed method when transferring knowledge (weights) from existing networks to our candidate architectures. For this purpose, we first define an existing (pre-trained) network to provide the weights for the modules of the candidate architectures.
In this work, we employ ResNet110 due to its high accuracy.
We highlight that since our candidate architectures are very shallow, we could employ shallower ResNets (e.g., ResNet56) and still avoid the restriction that the depth of the candidate architecture cannot exceed the depth of the network providing the weights (see Section~\ref{sec:proposed_method}).

Table~\ref{tab:comparison_scratch_ws_resnet} summarizes the results. Compared to training from scratch, our method with the weight transfer technique yields higher performance architectures in terms of depth, memory usage, number of FLOPs and parameters. 
The reason for different architectures when using training from scratch and the weight transfer technique (fine-tuning) is that the weights of the networks are different and influence the importance score directly. This leads to the insertion of modules on different stages throughout iterations.

Besides designing higher-performance architectures, an advantage of weight transfer is that it reduces the computational burden of training the architectures from scratch.
Specifically, this strategy reduces the average time for each iteration from $17$ to $3$ hours. We emphasize that the time for generating our architectures is faster than previous works, which require many days on several GPUs even when training for a few epochs (i.e., $20$)~\cite{Zoph:2018, Baker:2017, Real:2017}. For example, the approaches by Baker et al.~\cite{Baker:2017} and Real et al.~\cite{Real:2017} require $10$ days to discover competitive architectures.
%
It is worth mentioning that our method could be made even faster by training/fine-tuning architectures for only a few epochs, as suggested by previous works. On the other hand, this strategy can yield poor architectures, harming the search process~\cite{Dong:2020, Sciuto:2020}.
%

In summary, the proposed weight transfer technique speeds-up our NAS approach considerably, which enables searching architectures directly on large datasets. However, when using this technique the candidate architectures can take advantage of well-trained networks. Therefore, to make a fairer comparison with other NAS, unless stated otherwise, we are considering our method with training from scratch.

\noindent
\textbf{Comparison with human-designed architectures.} 
As we mentioned previously, human-designed architectures are generally composed of stages with uniform depth.
Our method, on the other hand, designs architectures by adjusting the depth for each stage based on its importance. To demonstrate that this process leads to more efficient and accurate networks, in this experiment, we compare our discovered architectures to their human-designed counterpart.

Table~\ref{tab:comparison_scratch_ws_resnet} compares our architectures with residual modules to existing residual networks~\cite{He:2016}. Compared to these networks, our architectures achieve superior performance in terms of the number of parameters, FLOPs, memory usage and accuracy. In particular, with one iteration our discovered architecture achieves comparable accuracy to ResNet110, having less than half of its computational cost.
Additionally, because our architectures are much shallower we achieve a significant improvement in inference time compared to ResNet110.
%
%

Following a recent trend~\cite{Lacoste:2019, Schwartz:2019, Strubell:2019}, we also measure the carbon emission for training architectures.
For this purpose, we use the Machine Learning Emissions Calculator~\cite{Lacoste:2019} and report the CO$_2$-equivalents (CO$_2$eq), which indicates the global-warming potential of various greenhouse gases as a single number.
According to Table~\ref{tab:comparison_scratch_ws_resnet}, our candidate architectures emit notably less carbon, even taking into account shallow versions of ResNet. Compared to ResNet110, our final architecture trained from scratch emits $41\%$ less CO$_2$. 
%
Observe that, from iteration one to five, our architectures have their carbon emission increased slightly whereas from ResNet44 to ResNet110 the increase is notably higher. This occurs because our architectures are computationally more efficient, leading to a considerably faster training stage.
%
Moreover, ResNets (as well as most human-designed architectures) have the same depth for all stages. Our architectures, on the other hand, had the depth of stages adjusted according to its importance, avoiding unnecessary growth in computational cost. 

Finally, the results in Table~\ref{tab:comparison_scratch_ws_resnet} show that adjusting depth on a stage-by-stage basis enables increasing capacity (reflected by accuracy) of networks without compromising their efficiency.
\begin{table}[!t]
	\centering
	\renewcommand{\arraystretch}{1.2}
	\caption{Performance of networks built with the proposed method and cell modules discovered by NAS.}
	\label{tab:comparison_nasnet}
	\begin{tabular}{cccccc}
		\hline
		Architecture & Depth & \begin{tabular}[c]{@{}c@{}}Param.\\ (Million)\end{tabular} & \begin{tabular}[c]{@{}c@{}}FLOPs\\ (Billion)\end{tabular} & \begin{tabular}[c]{@{}c@{}}Memory\\ (MB)\end{tabular} & \begin{tabular}[c]{@{}c@{}}Accuracy\\ CIFAR-10\end{tabular} \\ \hline
		NASNet169 & 169 & 2.3 & 2.7 & 71.26 & 94.34 \\
		Ours (it=1) & 109 & 1.3 & 1.8 & 45.20 & 94.55 \\ \hline
		NASNet205 & 205 & 2.8 & 3.2 & 86.37 & 94.37 \\
		Ours (it=3) & 133 & 1.5 & 2.2 & 56.96 & 94.63 \\ \hline
		NASNet241 & 241 & 3.3 & 3.8 & 101.47 & 94.51 \\
		Ours (it=5) & 181 & 2.3 & 2.9 & 78.87 & 94.74 \\ \hline
	\end{tabular}
\end{table}
\noindent
\textbf{Combination with other NAS approaches.} As we mentioned earlier, our method can employ modules discovered by other NAS approaches. We highlight that the NAS approaches that focus on discovering cells define the depth of stages uniformly (similar to human-designed architectures). In this experiment, we show that using these modules coupled with our strategy provides even better architectures.

Table~\ref{tab:comparison_nasnet} shows the results of our method applying cells as modules. Similar to residual modules, our architectures based on cells outperform those based on stages with uniform depth. Compared to the original NASNet ($m=6$ for all stages --- 241 layers deep) by Zoph~\cite{Zoph:2018}, with one iteration, our discovered architecture achieves superior accuracy having $60\%$, $52\%$ and $55\%$ fewer parameters, FLOPs and memory usage.

\noindent
\textbf{Comparison with state-of-the-art NAS.}
This experiment compares our method with state-of-the-art NAS approaches.

According to Table~\ref{tab:NAS}, our method is the more cost-effective NAS approach in terms of the number of evaluated models and amount of GPUs required.
Compared to Baker et al.~\cite{Baker:2017} and Real et al.~\cite{Real:2017}, our method designs competitive architectures by evaluating a significantly smaller number of models, enabling the proposed method to run in a few hours on a single GPU.
Specifically, our method evaluates one order of magnitude fewer models. This is because instead of analyzing a large pool of architectures like most approaches, we increment previous architectures iteratively while taking into account the importance of the components to be inserted. This advantage enables our method to scale to large datasets, while most NAS approaches might be prohibitive even when employing morphism and other optimization techniques~\cite{Ha:2017, Dong:2020}.
%
%
Compared to approaches that also evaluate a small number of models~\cite{Elsken:2018, Kandasamy:2018, Jin:2019}, our method achieves the best tradeoff between accuracy and number of GPUs required.

In summary, our method achieves competitive results using both residual and cell modules. 
When considering our best setting (cell modules with five iterations, see Table~\ref{tab:comparison_nasnet}), only the approaches by Elsken et al.~\cite{Elsken:2018} and Dong et al.~\cite{Dong:2019} obtain superior accuracy. 
We emphasize that our training process employs a simple SGD optimizer with standard data augmentation, while other NAS approaches employ sophisticated optimizers and regularization techniques (e.g., SGDR~\cite{Loshchilov:2017} and ScheduledDropPath~\cite{Zoph:2018}).
%
Although we could use these setups, they render it more difficult to identify which aspects actually lead to the improvement in NAS, as argued by Dong et al~\cite{Dong:2020}. Thus, we prefer to maintain the training as simple as possible.

Finally, our method built more parameter-efficient architectures even without considering the computational cost in the searching process. This occurs since most NAS approaches focus on discovering components of the architecture while keeping a uniform distribution of depth of the stages. 
Instead, our method adjusts this depth based on its importance leading to shallower, and hence more efficient, architectures.

%
%
%
\begin{table}[!t]
	\centering
	\renewcommand{\arraystretch}{1.2}
	\caption{Comparison with state-of-the-art NAS approaches. Results taken from previous works.}
	\label{tab:NAS}
	\begin{tabular}{ccccc}
		\hline
		Model & \begin{tabular}[c]{@{}c@{}}Evaluated \\ Models\end{tabular} & GPUs & \begin{tabular}[c]{@{}c@{}}Param.\\ (Million)\end{tabular} & \begin{tabular}[c]{@{}c@{}}Accuracy\\CIFAR-10 \end{tabular} \\ \hline
		Zoph et al.~\cite{Zoph:2018} & 20,000 & 800 & 2.5 & 94.51 \\
		Baker et al.~\cite{Baker:2017} & 1,500 & 10 & 11.1 & 93.08 \\	
    	Real et al.~\cite{Real:2017} & 1,000 & 250 & 5.4 & 94.60 \\
		Brock et al.~\cite{Brock:2018} & 300 & 1 & 4.6 & 94.47 \\
		Dong et al.~\cite{Dong:2019} & 240 & 1 & 2.5 & 96.25 \\ 
        Jin et al.~\cite{Jin:2019} & $\approx$60 & 1 & -- & 88.56
		\\
		Elsken et al.~\cite{Elsken:2018} & 40 & 5 & 19.7 & 94.80
		\\
    	Kandasamy et al.~\cite{Kandasamy:2018} & 10 & 4 & -- & 91.31 \\ \hline
		Ours (Res. modules)& 11 & 1 & 1.7 & 94.27 \\ 
		Ours (Cell modules) & 11 & 1 & 2.3 & 94.74 \\
		Ours (Ensemble) & -- & -- & 7.27 & 95.68 \\\hline
	\end{tabular}
\end{table}
\noindent
\textbf{Learning Architectures on Large Datasets.} 
Since our approach explores few candidate architectures, it is scalable to large datasets.
%
To reinforce this, we apply the proposed method to discover architectures on the large ImageNet ($224\times224$) dataset.
We use bottleneck residual blocks~\cite{He:2016} as modules and the weight transfer technique. Due to limitations for training NAS-based architectures, we do not consider cell modules.

Our final architecture obtained a top-5 accuracy of $90.23$, which is less than one percentage point inferior to the architecture by Dong et al.~\cite{Dong:2019}. Although it achieved a lower accuracy, an advantage of our method is that it can be applied on large datasets without requiring careful parameter setting since we are using the same parameters found on CIFAR-10. This advantage is desirable when no resources are available for tuning such parameters.
We highlight that the approach by Dong et al.~\cite{Dong:2019} fails to design networks directly on ImageNet, as it needs careful tuning and different hyper-parameters.

\noindent
\textbf{Generalization Ability.} 
An alternative to learning architectures on large datasets is to design them on small datasets and then transfer them to large datasets. The accuracy of the resulting architecture (trained from scratch) can be used to estimate its generalization ability (i.e., transferability), which is a desirable property in NAS~\cite{Zoph:2018}.
In this experiment, we assess this generalization ability of our architectures. For this purpose, we follow the same process by Zoph et al.~\cite{Zoph:2018}, which consists of taking an architecture found for CIFAR-10 and training it from scratch on other datasets.

Table~\ref{tab:large_datasets} shows the top-5 accuracy obtained on the Tiny ImageNet and ImageNet $32\times32$ datasets. 
For both datasets, when using residual modules, our architectures outperformed those based on stages with uniform depth. Specifically, our architecture obtained an accuracy up to $2.4$ p.p superior to ResNet110. 
With cell modules, our architecture and NASNet241 achieved similar results.
%
In summary, these results show that our architectures present high generalization ability.
\begin{table}[!t]
	\centering
	\renewcommand{\arraystretch}{1.2}
	\caption{Accuracy of networks transferred from a small dataset (i.e., CIFAR-10) to large datasets. The higher the accuracy the higher the generalization ability.}
	\label{tab:large_datasets}
	\begin{tabular}{ccc}
		\hline
		Architecture & TinyImageNet & ImageNet $32\times32$ \\ \hline
		ResNet110 & 69.94 & 68.89 \\
		Ours (Res. modules) & 72.34 & 70.09 \\ \hline
		NASNet241 & 79.20 & 80.67 \\
		Ours (Cell modules) & 79.23 & 79.39 \\ \hline
	\end{tabular}
\end{table}
\noindent
\textbf{Ensemble of Architectures.} Motivated by the fact that an ensemble of candidate architectures can obtain better accuracy than the final architecture alone~\cite{Elsken:2018}, our last experiment shows the performance of our method employing this strategy.

Our ensemble is composed of the candidate architectures presented in Tables~\ref{tab:comparison_scratch_ws_resnet} and~\ref{tab:comparison_nasnet}, achieving an accuracy of $95.68$ with $7.27$ million parameters. Compared to the ensemble of Elsken et al.~\cite{Elsken:2018}, which obtains an accuracy of $95.60$ with $88$ million parameters, our ensemble is marginally more accurate having $12\times$ fewer parameters. In particular, our ensemble is more parameter-efficient even when compared to a single architecture of Elsken et al.~\cite{Elsken:2018} (see Table~\ref{tab:NAS}). 
%
\section{Conclusions}\label{sec:conclusions}
In this work, we proposed a simple, effective, and efficient approach to discover convolutional architectures. Our method designs candidate architectures by adjusting the depth of each stage based on its importance.
%
%
Compared to previous NAS approaches, our method is significantly more efficient as it evaluates one order of magnitude fewer models. 
Our discovered architectures are on par with the state-of-the-art and present high generalization ability, since they can be learned on a small dataset and successfully transferred to large datasets.
\section*{Acknowledgments}
The authors would like to thank the National Council for Scientific and Technological Development -- CNPq (Grants~140082/2017-4, ~438629/2018-3 and~309953/2019-7) and the Minas Gerais Research Foundation -- FAPEMIG (Grants~APQ-00567-14 and~PPM-00540-17.

%
\IEEEpeerreviewmaketitle

%
%

\bibliographystyle{ieeetran}
\bibliography{refs}

\begin{thebibliography}{10}
\providecommand{\url}[1]{#1}
\csname url@samestyle\endcsname
\providecommand{\newblock}{\relax}
\providecommand{\bibinfo}[2]{#2}
\providecommand{\BIBentrySTDinterwordspacing}{\spaceskip=0pt\relax}
\providecommand{\BIBentryALTinterwordstretchfactor}{4}
\providecommand{\BIBentryALTinterwordspacing}{\spaceskip=\fontdimen2\font plus
\BIBentryALTinterwordstretchfactor\fontdimen3\font minus
  \fontdimen4\font\relax}
\providecommand{\BIBforeignlanguage}[2]{{%
\expandafter\ifx\csname l@#1\endcsname\relax
\typeout{** WARNING: IEEEtran.bst: No hyphenation pattern has been}%
\typeout{** loaded for the language `#1'. Using the pattern for}%
\typeout{** the default language instead.}%
\else
\language=\csname l@#1\endcsname
\fi
#2}}
\providecommand{\BIBdecl}{\relax}
\BIBdecl

\bibitem{He:2016}
K.~He, X.~Zhang, S.~Ren, and J.~Sun, ``Deep residual learning for image
  recognition,'' in \emph{CVPR}, 2016.

\bibitem{Tan:2019:ICML}
M.~Tan and Q.~V. Le, ``Efficientnet: Rethinking model scaling for convolutional
  neural networks,'' in \emph{ICML}, 2019.

\bibitem{Strubell:2019}
E.~Strubell, A.~Ganesh, and A.~McCallum, ``Energy and policy considerations for
  deep learning in {NLP},'' in \emph{ACL}, 2019.

\bibitem{Huang:2017}
G.~Huang, Z.~Liu, L.~van~der Maaten, and K.~Q. Weinberger, ``Densely connected
  convolutional networks,'' in \emph{CVPR}, 2017.

\bibitem{He:2015}
K.~He, X.~Zhang, S.~Ren, and J.~Sun, ``Delving deep into rectifiers: Surpassing
  human-level performance on imagenet classification,'' in \emph{ICCV}, 2015.

\bibitem{Howard:2017}
A.~G. Howard, M.~Zhu, B.~Chen, D.~Kalenichenko, W.~Wang, T.~Weyand,
  M.~Andreetto, and H.~Adam, ``Mobilenets: Efficient convolutional neural
  networks for mobile vision applications,'' \emph{arXiv}, 2017.

\bibitem{Sandler:2018}
M.~Sandler, A.~G. Howard, M.~Zhu, A.~Zhmoginov, and L.~Chen, ``Mobilenetv2:
  Inverted residuals and linear bottlenecks,'' in \emph{CVPR}, 2018.

\bibitem{Wang:2018}
R.~J. Wang, X.~Li, and C.~X. Ling, ``Pelee: A real-time object detection system
  on mobile devices,'' in \emph{NeurIPS}, 2018.

\bibitem{Gordon:2018}
A.~Gordon, E.~Eban, O.~Nachum, B.~Chen, H.~Wu, T.~Yang, and E.~Choi,
  ``Morphnet: Fast {\&} simple resource-constrained structure learning of deep
  networks,'' in \emph{CVPR}, 2018.

\bibitem{Jin:2019}
H.~Jin, Q.~Song, and X.~Hu, ``Auto-keras: An efficient neural architecture
  search system,'' in \emph{SIGKDD}, 2019.

\bibitem{Dong:2019}
X.~Dong and Y.~Yang, ``Searching for a robust neural architecture in four {GPU}
  hours,'' in \emph{CVPR}, 2019.

\bibitem{Real:2017}
E.~Real, S.~Moore, A.~Selle, S.~Saxena, Y.~L. Suematsu, J.~Tan, Q.~V. Le, and
  A.~Kurakin, ``Large-scale evolution of image classifiers,'' in \emph{ICML},
  2017.

\bibitem{Chen:2019}
C.~R. Chen, Q.~Fan, N.~Mallinar, T.~Sercu, and R.~S. Feris, ``Big-little net:
  An efficient multi-scale feature representation for visual and speech
  recognition,'' in \emph{ICLR}, 2019.

\bibitem{Huang:2018}
Q.~Huang, S.~K. Zhou, S.~You, and U.~Neumann, ``Learning to prune filters in
  convolutional neural networks,'' in \emph{WACV}, 2018.

\bibitem{Wang:2018:ECCV}
X.~Wang, F.~Yu, Z.~Dou, T.~Darrell, and J.~E. Gonzalez, ``Skipnet: Learning
  dynamic routing in convolutional networks,'' in \emph{ECCV}, 2018.

\bibitem{Greff:2017}
K.~Greff, R.~K. Srivastava, and J.~Schmidhuber, ``Highway and residual networks
  learn unrolled iterative estimation,'' in \emph{ICLR}, 2017.

\bibitem{Geladi:1986}
P.~Geladi and B.~Kowalski, ``Partial least-squares regression: a tutorial,''
  \emph{Analytica Chimica Acta}, vol. 185, 1986.

\bibitem{Roffo:ICCV}
G.~Roffo, S.~Melzi, and M.~Cristani, ``Infinite feature selection,'' in
  \emph{ICCV}, 2015.

\bibitem{Roffo:2017}
G.~Roffo, S.~Melzi, U.~Castellani, and A.~Vinciarelli, ``Infinite latent
  feature selection: {A} probabilistic latent graph-based ranking approach,''
  in \emph{ICCV}, 2017.

\bibitem{CVPR:2017}
S.~Xie, R.~B. Girshick, P.~Doll{\'{a}}r, Z.~Tu, and K.~He, ``Aggregated
  residual transformations for deep neural networks,'' in \emph{CVPR}, 2017.

\bibitem{Veit:2016}
A.~Veit, M.~J. Wilber, and S.~J. Belongie, ``Residual networks behave like
  ensembles of relatively shallow networks,'' in \emph{NeurIPS}, 2016.

\bibitem{krizhevsky2009}
A.~Krizhevsky and G.~Hinton, ``Learning multiple layers of features from tiny
  images,'' University of Toronto, Tech. Rep., 2009.

\bibitem{Le:2015}
Y.~Le and Y.~Xuan, ``Tiny imagenet visual recognition challenge,'' \emph{CS
  231N}, 2015.

\bibitem{ilsvrc15}
O.~Russakovsky, J.~Deng, H.~Su, J.~Krause, S.~Satheesh, S.~Ma, Z.~Huang,
  A.~Karpathy, A.~Khosla, M.~Bernstein, A.~C. Berg, and L.~Fei-Fei, ``Imagenet
  large scale visual recognition challenge,'' \emph{IJCV}, vol. 115, no.~3, pp.
  211--252, 2015.

\bibitem{Chrabaszcz:2017}
P.~Chrabaszcz, I.~Loshchilov, and F.~Hutter, ``A downsampled variant of
  imagenet as an alternative to the {CIFAR} datasets,'' \emph{arXiv}, 2017.

\bibitem{Lacoste:2019}
A.~Lacoste, A.~Luccioni, V.~Schmidt, and T.~Dandres, ``Quantifying the carbon
  emissions of machine learning,'' in \emph{NeurIPS}, 2019.

\bibitem{Schwartz:2019}
R.~Schwartz, J.~Dodge, N.~A. Smith, and O.~Etzioni, ``Green {AI},''
  \emph{arXiv}, 2019.

\bibitem{Baker:2017}
B.~Baker, O.~Gupta, N.~Naik, and R.~Raskar, ``Designing neural network
  architectures using reinforcement learning,'' in \emph{ICLR}, 2017.

\bibitem{Zoph:2018}
B.~Zoph, V.~Vasudevan, J.~Shlens, and Q.~V. Le, ``Learning transferable
  architectures for scalable image recognition,'' in \emph{CVPR}, 2018.

\bibitem{Wu:2018}
Z.~Wu, T.~Nagarajan, A.~Kumar, S.~Rennie, L.~S. Davis, K.~Grauman, and R.~S.
  Feris, ``Blockdrop: Dynamic inference paths in residual networks,'' in
  \emph{CVPR}, 2018.

\bibitem{Dong:2020}
X.~Dong and Y.~Yang, ``Nas-bench-201: Extending the scope of reproducible
  neural architecture search,'' in \emph{ICLR}, 2020.

\bibitem{Elsken:2018}
T.~Elsken, J.~H. Metzen, and F.~Hutter, ``Simple and efficient architecture
  search for convolutional neural networks,'' in \emph{ICLR}, 2018.

\bibitem{Kandasamy:2018}
K.~Kandasamy, W.~Neiswanger, J.~Schneider, B.~P{\'{o}}czos, and E.~P. Xing,
  ``Neural architecture search with bayesian optimisation and optimal
  transport,'' in \emph{NeurIPS}, 2018.

\bibitem{Chen:2016}
T.~Chen, I.~J. Goodfellow, and J.~Shlens, ``Net2net: Accelerating learning via
  knowledge transfer,'' in \emph{ICLR}, 2016.

\bibitem{Cai:2018}
H.~Cai, T.~Chen, W.~Zhang, Y.~Yu, and J.~Wang, ``Efficient architecture search
  by network transformation,'' in \emph{AAAI}, 2018.

\bibitem{Sciuto:2020}
C.~Sciuto, K.~Yu, M.~Jaggi, C.~Musat, and M.~Salzmann, ``Evaluating the search
  phase of neural architecture search,'' in \emph{ICLR}, 2020.

\bibitem{Jordao:2020}
A.~{Jordao}, M.~{Lie}, and W.~R. {Schwartz}, ``Discriminative layer pruning for
  convolutional neural networks,'' \emph{IEEE Journal of Selected Topics in
  Signal Processing}, 2020.

\bibitem{Mehmood:2012}
T.~Mehmood, K.~H. Liland, L.~Snipen, and S.~Saebo, ``A review of variable
  selection methods in partial least squares regression,'' \emph{Chemometrics
  and Intelligent Laboratory Systems}, 2012.

\bibitem{Yu:2018}
R.~Yu, A.~Li, C.~Chen, J.~Lai, V.~I. Morariu, X.~Han, M.~Gao, C.~Lin, and L.~S.
  Davis, ``{NISP:} pruning networks using neuron importance score
  propagation,'' in \emph{CVPR}, 2018.

\bibitem{Brock:2018}
A.~Brock, T.~Lim, J.~M. Ritchie, and N.~Weston, ``{SMASH:} one-shot model
  architecture search through hypernetworks,'' in \emph{ICLR}, 2018.

\bibitem{Ha:2017}
D.~Ha, A.~M. Dai, and Q.~V. Le, ``Hypernetworks,'' in \emph{ICLR}, 2017.

\bibitem{Loshchilov:2017}
I.~Loshchilov and F.~Hutter, ``{SGDR:} stochastic gradient descent with warm
  restarts,'' in \emph{ICLR}, 2017.

\end{thebibliography}

\end{document}